\documentclass[11pt,a4paper]{article}
\usepackage{authblk}
\usepackage{times}
\usepackage{latexsym}
\usepackage{booktabs}
\usepackage[shortlabels]{enumitem}
\usepackage{multirow}
\usepackage{amsmath}
\usepackage{amssymb}
\usepackage{amsfonts}
\usepackage{graphicx}
\usepackage{float}
\usepackage{subfig}
\usepackage{url}
\usepackage{todonotes}
\usepackage[hyperref]{naaclhlt2019}

\aclfinalcopy 

\title{On the Feasibility of Automated Detection of Allusive Text Reuse}
\author[1]{\textbf{Enrique Manjavacas}}
\author[2]{\textbf{Brian Long}}
\author[1]{\textbf{Mike Kestemont}}
\affil[1]{Computational Linguistics and Psycholinguistics Research Center\\University of Antwerp\\Belgium}
\affil[2]{University of Notre Dame\\United States of America\protect\\\tt{\{firstname,lastname\}@uantwerpen.be}\\\tt{blong2@alumni.nd.edu}}

\date{}

\begin{document}

\maketitle

\begin{abstract}
The detection of allusive text reuse is particularly challenging due to the sparse evidence on which allusive references rely---commonly based on none or very few shared words. %
Arguably, lexical semantics can be resorted to since uncovering semantic relations between words has the potential to increase the support underlying the allusion and alleviate the lexical sparsity. A further obstacle is the lack of evaluation benchmark corpora, largely due to the highly interpretative character of the annotation process. In the present paper, we aim to elucidate the feasibility of automated allusion detection. We approach the matter from an Information Retrieval perspective in which referencing texts act as queries and referenced texts as relevant documents to be retrieved, and estimate the difficulty of benchmark corpus compilation by a novel inter-annotator agreement study on query segmentation. Furthermore, we investigate to what extent the integration of lexical semantic information derived from distributional models and ontologies can aid retrieving cases of allusive reuse. The results show that (i) despite low agreement scores, using manual queries considerably improves retrieval performance with respect to a windowing approach, and that (ii) retrieval performance can be moderately boosted with distributional semantics.

\end{abstract}

\section{Introduction}
In the 20th century, intertextuality emerged as an influential concept in literary criticism. Originally developed by French deconstructionist theorists, such as Kristeva and Barthes, the term broadly refers to the phenomenon where texts integrate (fragments of) other texts or allude to them \cite{Orr2003Intertextuality:Contexts}. In the minds of both authors and readers, intertexts can establish meaningful connections between works, evoking particular stylistic effects and interpretations of a text. Existing categorizations \cite{Bamman2008TheAllusion,mellerin:halshs-01543349,buchler2013informationstechnische,hohl2010key} emphasize the broad spectrum of intertexts, which can range from direct quotations, over paraphrased passages to highly subtle allusions. 

With the emergence of computational methods in literary studies over the past decades, intertextuality has often been presented as a promising application, helping scholars identifying potential intertextual links that had previously gone unnoticed. Much progress has been made in this area and a number of highly useful tools are now available---e.g. Tracer \cite{buchler2013informationstechnische} or Tesserae \cite{coffee2012tesserae}. This paper, however, aims to contribute to a number of open issues that still present significant challenges to the further development of the field.

\begin{figure}
    \centering
    \begin{quote}
        \textbf{Reference} (Vulgata, Ep 3,19) ``scire etiam supereminentem scientiae caritatem Christi ut impleamini in omnem plenitudinem Dei''

        ``and to know the love (caritas) of Christ that is beyond knowledge, such that you'd be filled with all fullness of God''

        \textbf{Reuse} (Bernard, Sermo 8, 7.l) ``Osculum plane dilectionis et pacis, \textit{sed dilectio illa supereminet omni \textbf{scientiae}}, et pax illa omnem sensum exsuperat''

        ``It is a kiss of love and peace, but of that kind of love (dilectio) that is beyond any knowledge, and of that kind of peace that surpasses all senses.''
    \end{quote}

    \caption{Examples of allusive text reuse from the dataset underlying the present study.}
    \label{fig:quote}
\end{figure}

Most scholarship continues to focus on the detection of relatively literal instances of so-called `text reuse', as intertextuality is commonly -- and somewhat restrictively -- referred to in the field. Such instances are relatively unambiguous and unproblematic to detect using n-gram matching, fingerprinting and string alignment algorithms. Much less research has been devoted to the detection of fuzzier instances of text reuse holding between passages that lack a significant lexical correspondence. This situation is aggravated by the severe lack of openly available benchmark datasets. 
An additional hindrance is that the establishment of intertextual links is to a high degree subjective -- both regarding the existence of particular intertextual links and the exact scope of the correspondence in both fragments. Studies of inter-annotator agreement are surprisingly rare in the field, which might be partially due to to the fact that existing agreement metrics are hard to port to this problem.

\paragraph{Contributions} In this paper, we report on an empirical feasibility study, focusing on the annotation and automated detection of allusive text reuse. We focus on biblical intertext in the works of Bernard of Clairvaux (1090--1153), an influential medieval writer known for his pervasive references to the Bible. The paper has two main parts. In the first part, we formulate an adaptation of Fleiss's $\kappa$ that allows us to quantitatively estimate and discuss the level of inter-annotator agreement concerning the span of the intertexts. While annotators show considerably low levels of agreement, We show that manual segmentation has nevertheless a big impact on the automatic retrieval of allusive reuse. In the second part, we offer an evaluation of current Information Retrieval (IR) techniques for allusive text reuse detection. We confirm that semantic retrieval models based on word and sentence embeddings do not present advantages over hand-crafted scoring functions from previous studies, and that both are outperformed by conventional retrieval models based on TfIdf. Finally, we show how a recently introduced technique, soft cosine, allows us to combine lexical and semantic information to obtain significant improvements over any other considered model.

\section{Related Work}
Previous research on text reuse detection in literary texts has extensively explored methods such as n-gram matching \cite{Buchler2014TowardsDetection} and sequence alignment algorithms \cite{Lee2007ATexts,Smith2014DetectingReuse}. In such approaches, fuzzier forms of intertextual links are accounted for through the use of edit distance comparisons or the inclusion of abstract linguistic information such as word lemmata or part-of-speech tags, and lexical semantic relationships extracted from WordNet. More recently, researchers have started to explore techniques from the field of distributional semantics in order to capture allusive text reuse. \citet{Scheirer2016TheMeaning}, for instance, have applied latent-semantic indexing (LSI) to find semantic connections and evaluated such method on a set of 35 allusive references to Vergil's \emph{Aeneis} in the first book of Lucan's \emph{Civil War}.

Previous research in the field of text reuse has also focused on the more specific problem of finding allusive references. One of the first studies \cite{Bamman2008TheAllusion} looked at allusion detection in literary text using an IR approach exploiting textual features at a diversity of levels (including morphology and syntax) but collected only qualitative evidence on the efficiency of such approach. More ambitiously, \citet{Bamman2009DiscoveringTexts} approached the task of finding allusive references across texts in different languages using string alignment algorithms from machine translation. Besides the afore-mentioned work by \citet{Scheirer2016TheMeaning}, the work by \citet{Moritz2016Non-literalReuse} is highly related to the present study, since the authors also worked on allusive reuse from the Bible in the works of Bernard. In their work, the authors focused on modeling text reuse patterns based on a set of transformation rules defined over string case, lemmata, POS tags and synset relationships: (syno-/hypo-/co-hypo-)nymy. More recently, \citet{moritz2018method} conducted a quantitative comparison of such transformation rules with paraphrase detection methods on the task of predicting paraphrase relation between text pairs but do not evaluate the method in an IR setup.


\section{Dataset}

The basis for the present study stems from the BiblIndex project \citep{mellerin:halshs-01543349}, which aims to index biblical references found in Christian literature.\footnote{
    \href{http://www.biblindex.mom.fr/}{http://www.biblindex.mom.fr/}
} More specifically, we use a subset of manually identified biblical references from Bernard of Clairvaux which was kindly shared with us by Laurence Mellerin. The provided data consists of 85 Sermons, totalling 199,508 words. The data came already tokenized and lemmatized. Bible references were tagged with a URL mapping to the corresponding Bible verse from the Vulgata edition of the medieval Bible in the online BiblIndex database. We extracted the online text of the Vulgata and used the URLs to match references in Bernard with the corresponding Bible verses. Since the online BiblIndex database does not provide lemmatized text, we applied an state-of-the-art lemmatizer for Medieval Latin \cite{manjavacas2019improving} to obtain a lemmatized version of the Vulgata. The resulting corpus data comprises a total of 34,835 verses totalling 586,285 tokens and amounting to a vocabulary size of 46,025 token types.

\begin{figure}[ht]
    \centering
    \includegraphics[width=0.85\linewidth]{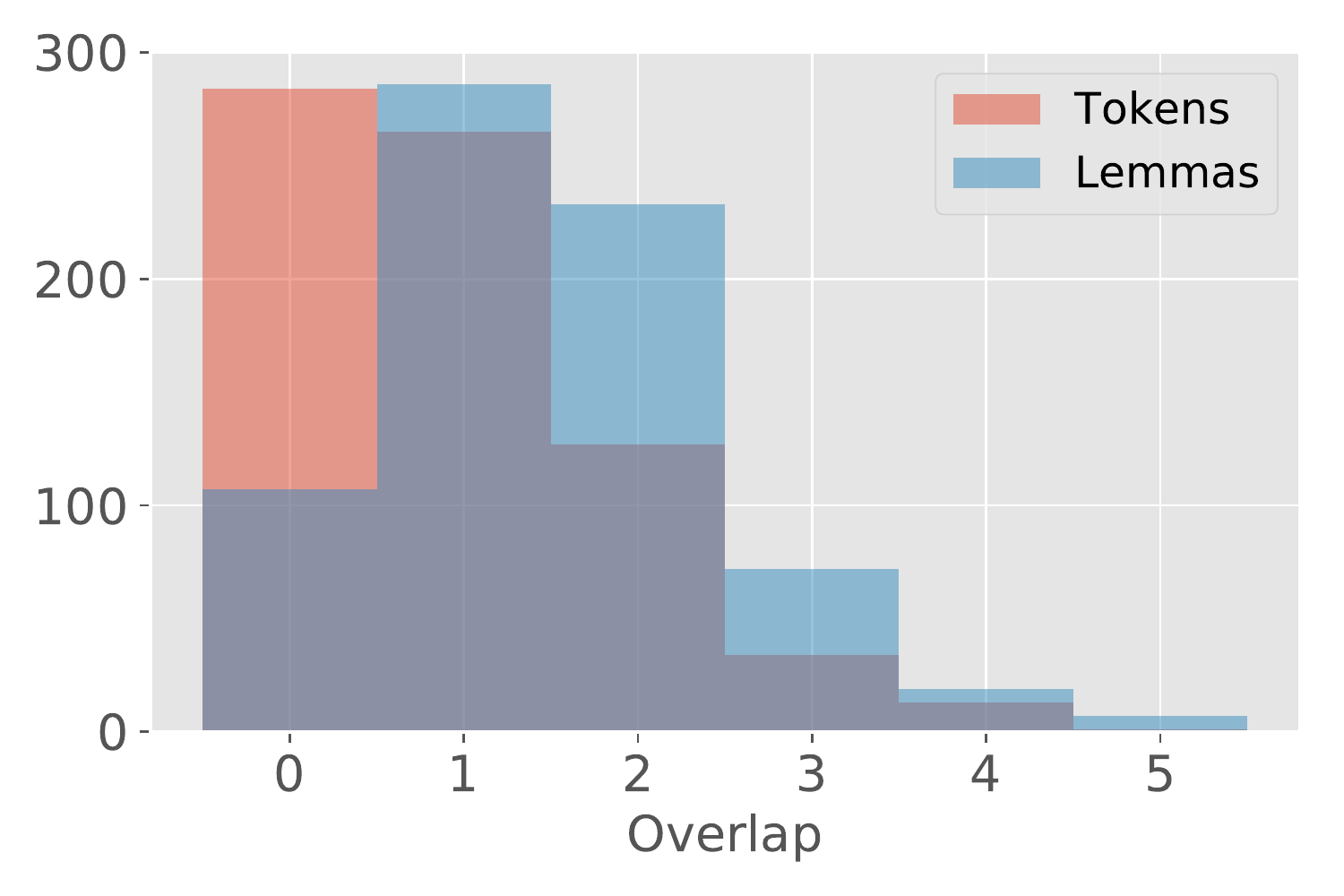}
    \caption{Histogram of token and lemma overlap between annotated queries and their Biblical references}
    \label{fig:overlap}
\end{figure}

BiblIndex distinguishes three types of references: quotation, mention and allusion. While the links in the first two types are in their vast majority exact or near-exact lexical matches, the latter type comprises mostly references that fall into what is commonly known as allusive text reuse. Although our focus lies on the allusive category, Table \ref{tab:stats} displays statistics about all these types in order to appreciate the characteristics of the task. As shown in Table \ref{tab:stats} (last row), allusions are characterized by low Jaccard coefficients -- in set-theoretical terms, the ratio of the intersection over the union of the sets of words of both passages. On average, annotated allusions share 6\% of the word forms with their targets and 12\% of the lemmata. In comparison, mentions and quotations have 25\% or more tokens and 30\% or more lemmata in common. The full distribution of token and lemma overlap for allusions shown in Fig.~\ref{fig:overlap} indicates that more than 500 (~65\%) instances have at most 1 token in common; about more than 400 (~50\%) share at most 1 lemma.

\begin{table*}[ht]
    \centering

     \begin{tabular}{cccccc}
        & Jaccard(token) & Jaccard(lemma) & Source length & Ref length & Count \\
        \midrule
        Quotation & 0.37 ($\pm$ 0.23) & 0.37 ($\pm$ 0.22) & 6.69 ($\pm$ 4.55) & 15.12 ($\pm$ 5.99) & 1768 \\
        Mention & 0.26 ($\pm$ 0.18) & 0.31 ($\pm$ 0.18) & 7.47 ($\pm$ 5.52) & 16.24 ($\pm$ 6.20) & 3150 \\
        Allusion & 0.02 ($\pm$ 0.04) & 0.04 ($\pm$ 0.05) & 1.10 ($\pm$ 0.85) & 17.22 ($\pm$ 6.58) & 876  \\
        \midrule
        Allusion (post) & 0.06 ($\pm$ 0.07) & 0.13 ($\pm$ 0.1) & 6.86 ($\pm$ 4.83) &  & 729 \\
        \bottomrule
    \end{tabular}
    \caption{
        Full dataset statistics for all link types originally provided by the editors. Last row shows statistics for allusive references in Bernard post annotation. We show Jaccard coefficients for original and lemmatized sentences, text lengths and instance counts. 
    }
    \label{tab:stats}
\end{table*}

\section{Annotation}
Conventional systems in text reuse detection typically work by segmenting texts into consecutive, equal-length chunks of texts, which are then used as queries to find cross-document matches. For (semi-)literal cases of reuse, this matching procedure yields good results and overlapping or adjacent matches can be easily merged into longer units of reuse. For allusive text reuse, such an approach seems unfeasible at the current stage, partially because the definition of the relevant query units is much harder to establish. As shown in Table \ref{tab:stats}, the annotated allusive references are mere `anchors', consisting of single words or single multi-word expressions that cannot be easily used as queries. This is in agreement with pragmatic editorial conventions, which favour uncompromising signposting of references at anchor words over establishing particular decisions on the scope of the reference. However, from the point of view of the evaluation of IR systems, the provided editorial anchors must be turned into fully-fleshed, neatly delineated queries. In order to accomplish this, we have conducted an annotation experiment, which we will describe next.

\subsection{Full dataset annotation}
The aim of the annotation was to determine the scope of a biblical reference identified by the editors in text by Bernard. From an IR perspective, the annotation task consists of delineating the appropriate input query, given the anchor word in the source text and the corresponding Bible verse. An example annotation is shown in Fig.~\ref{fig:quote} where the anchor word provided by the editors is ``scientiae'' and the corresponding annotated query spans the subclause ``sed dilection illa supereminet omni scientiae''. Naturally, such references not always correspond to full sentences and often go over sentence boundaries.


The dataset was distributed evenly across 4 annotators, who worked independently through a custom-built interface. All annotators were proficient readers of Medieval Latin with expertise ranging from graduate student to professor. The annotators were familiar with the text reuse detection task and were given explicit instructions that can be summarized as follows: given a previously identified allusion between the Bernardine passage surrounding an anchor word, on the one hand, and a specific Bible verse on the other hand, annotate the \emph{minimal textual span} in the Bernardine passage that is \emph{maximally allusive} to the Bible verse. For the sake of simplicity, the interface only allowed continuous annotation spans and the annotated span had to include the pre-identified anchor token. Of a total of 876 initial instances, we discarded 147 cases in which annotators expressed doubts on the existence of the alleged reference or could not precisely decide the span. This decision was taken in order to ensure a high quality in the resulting benchmark data.


\subsection{Inter-annotator agreement experiment}
Determining the scope of an allusive reference is a relevant task for two reasons. Firstly, we expect this task to be reader-dependent, and thus highly subjective, given the minimal lexical overlap between the source and target passage. Measuring the agreement between annotators sheds new light on the overall feasibility of the task. Secondly, the resulting annotations allow us to critically evaluate the performance of existing retrieval methods under near-perfect segmentation conditions: if the correct source query is given, what is the performance of existing methods when attempting to retrieve the correct Bible verse in the target data?

\paragraph{Measuring inter-annotator agreement}
Inter-annotator agreement coefficients such as Fleiss's $\kappa$ and Krippendorff's $\alpha$ are typically defined in terms of labels assigned to items in a multi-class classification setup \cite{Artstein2008Inter-CoderLinguistics}. In the present case, however, the annotation involves making a decision on the span of words surrounding an anchor word that better captures the allusion and it is unclear how to quantify the variation in annotation performance. A na{\"i}ve approach defined in terms of number of overlaping words has a number of undesirable issues. For example, since the annotations are centered around the anchor word, a relatively high amount of overlap is to be expected for short annotations. Moreover, disagreements over otherwise largely agreeing long spans should weigh in less than disagreements over otherwise largely agreeing small spans. Additionally, it is unclear how to quantify the rate of agreement expected under chance-level annotation, a quantity that needs to be corrected for in order to to obtain reliable and non-inflated inter-annotator agreement coefficients \cite{artstein2017inter}. We have found that an extension of the Jaccard coefficient defined over sequences can help adapt Fleiss's $\kappa$ to our case and tackle such issues.

Given any pair of span annotations, $s$ and $t$, we can define overlap in a similar way to the Jaccard index, as the intersection (i.e. the Longest Common Substring) over the union (i.e. the total number of selected tokens by both annotators):

\begin{equation}
    O=\frac{LCS(s,t)}{|s| + |t| - LCS(s,t)}
\end{equation}

Interestingly, this quantity can be decomposed into an agreement $A(s,t) = LCS(s,t)$ (number of tokens in common) and a disagreement score $D(s,t)=|s|+|t|-2 \cdot LCS(s, t)$ (number of tokens not shared with the other annotator): 

\begin{equation}\label{eq:o_o}
    O = \frac{A}{A+D}
\end{equation}

The advantage of this reformulation is that it lets us see more easily how $O$ is bounded between 0 and 1, and also that it gives us a way of computing the expected overlap score $O_e$ by aggregating dataset-level $A$ and $D$ scores: $O_e=A_e/(A_e+D_e)$, with

\begin{equation}
     A_e=\frac{\sum_{s,t} A(s,t)}{|s,t|}; D_w=\frac{\sum_{s,t} D(s,t)}{|s,t|}    
\end{equation}

where $|s,t|$ refers to the number of unordered annotation pairs in the dataset\footnote{
    Such quantity is defined by $N k (k-1)/2$, where $N$ is the number of annotations and $k$ the number of annotators.
}. $O_e$ can be thus interpreted as the expected overlap between two arbitrary annotators. The final inter-annotator agreement score is defined following Fleiss's:
\begin{equation}
    \kappa = \frac{O_o-O_e}{1-O_e}
\end{equation}

where $O_o$ refers to the dataset average of Eq.~\ref{eq:o_o}.


\begin{figure*}[ht!]
    \centering
        \subfloat[][Observed Overlap]{\includegraphics[width=0.42\linewidth]{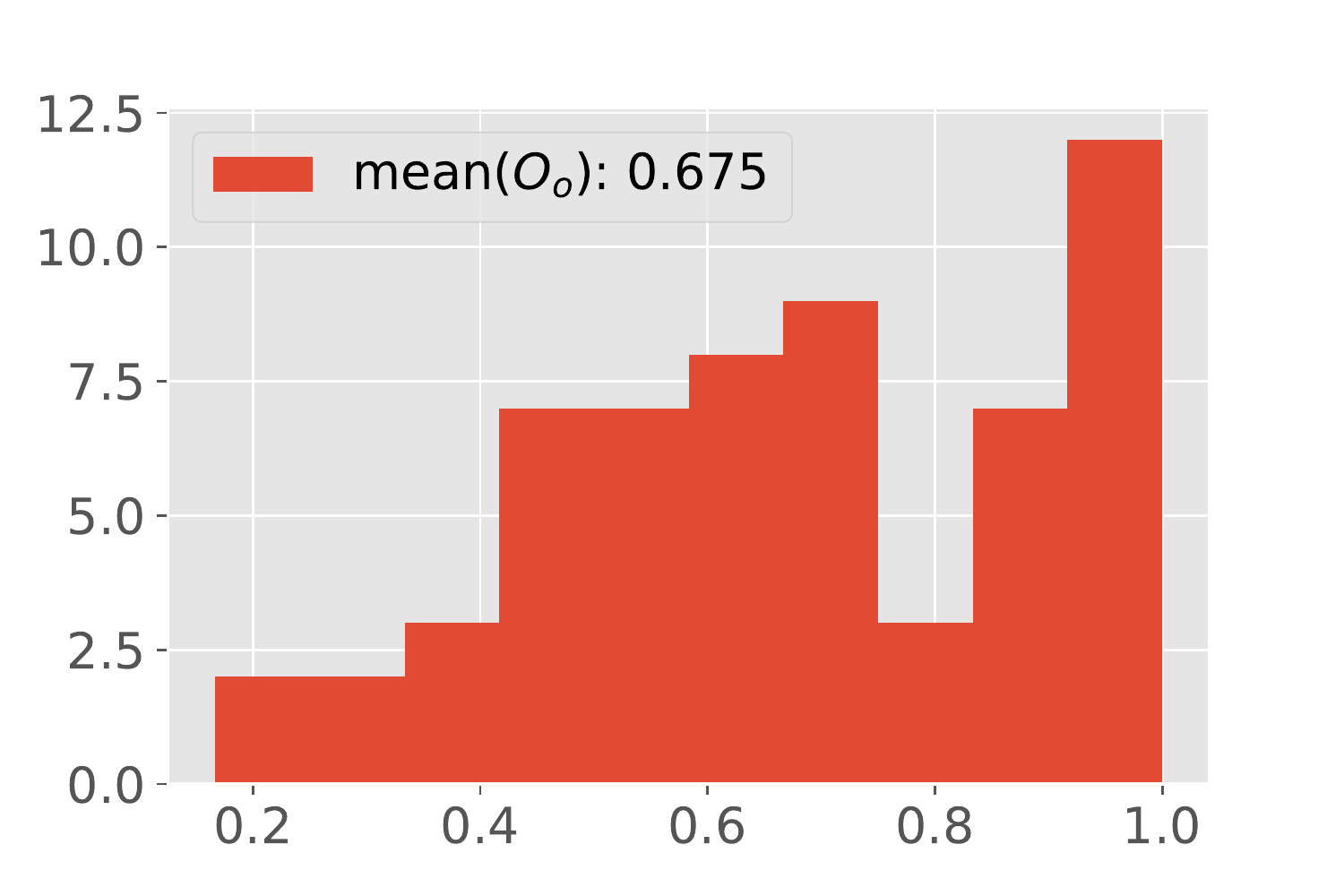}}%
        \subfloat[][Cumulative Overlap]{\includegraphics[width=0.42\linewidth]{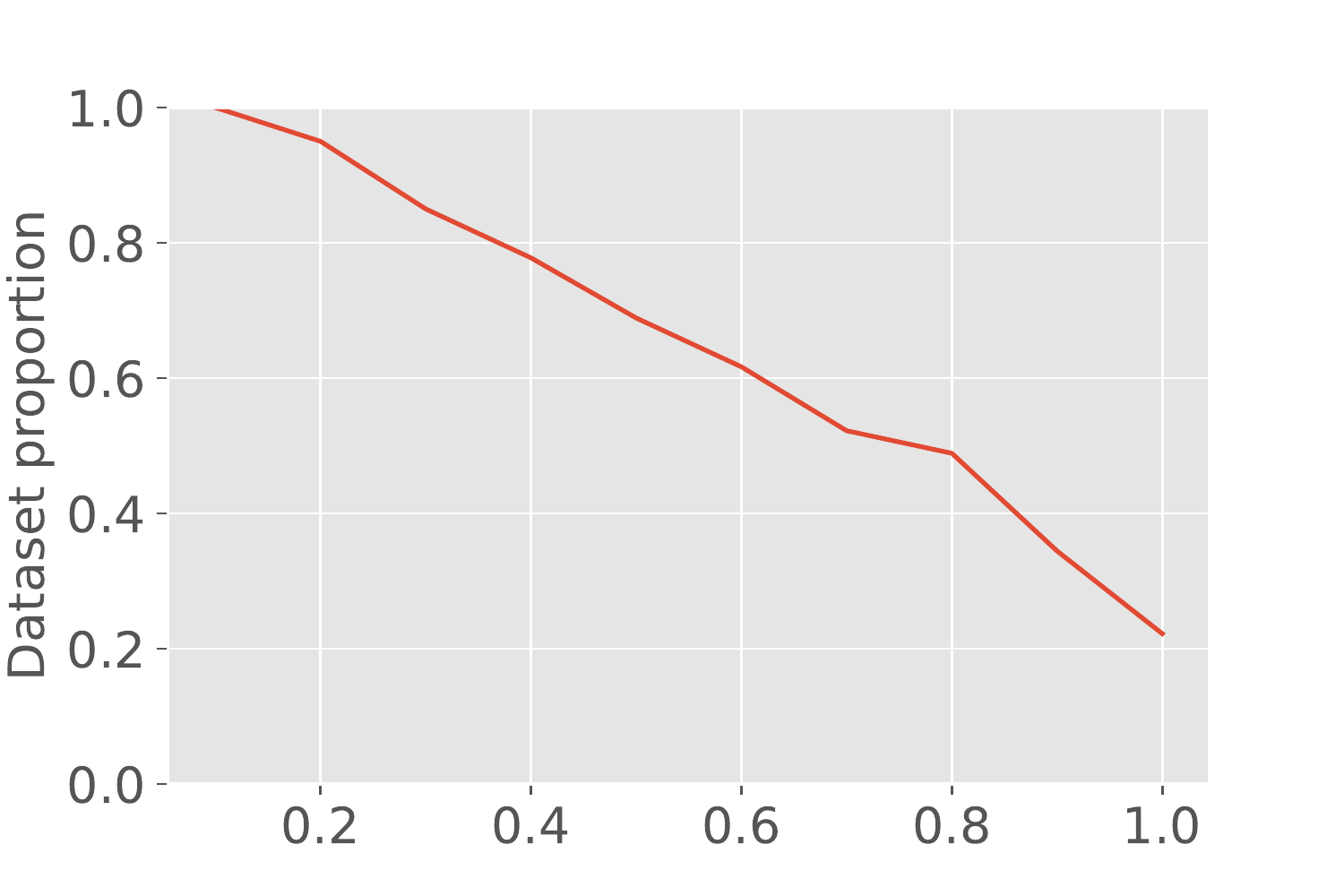}}%
    \caption{Observed overlap in the inter-annotator agreement experiments. On the left (a), we see the full histogram of $O_o$ in the dataset ($N=60$). On the right (b), we see the cumulative plot. We observe two modes in the histogram, perhaps indicating a qualitative difference in the dataset. One with high overlap scores close to 1.0 and another one at around 0.6 (close to the overall overlap mean).}
    \label{fig:agreement}
\end{figure*}

\paragraph{Inter-annotator agreement results and discussion} In order to estimate $\kappa$ for our dataset, we extracted a random sample of 60 instances which were thoroughly annotated by 3 of the annotators. We obtain a $\kappa = 0.22$, which compares unfavorably with respect to commonly assumed reliability ranges. For example, values in the range $\kappa \in (0.67, 0.8)$ are considered fair agreement \cite{schutze2008introduction}. While our result remains hard to assess in the absence of comparable work, it is low enough to cast doubts over the feasibility of the task, which is in fact rarely explicitly questioned. The annotators informally reported that, against their expectations, the task was not straightforward and required a considerable level of concentration and interpretation. Such situation may be due to particularities of Bernard's usage of biblical language. Besides conventional, direct allusions, Bernard is also known for pointed use of single, significant allusive words, which are hard to isolate. Still it should be noted that in some instances inter-annotator agreement was high and, as Fig.~\ref{fig:agreement}(b) shows, in 22\% of all pairwise comparisons even perfect. This suggests that there exist clear differences at the level of individual allusions. We now turn to the question how well current retrieval approaches perform, given manually segmented queries.

\section{Retrieval Experiments}

Given the small amounts of lexical overlap in the allusive text reuse datasets (c.f. Table~\ref{tab:stats}), we aim to investigate and quantify to which extent semantic information can help improving retrieval of allusive references. For this reason, we look into 3 types of models. First, we look at purely lexical-based approaches. Secondly, approaches based on distributional semantics and, in particular, retrieval approaches that utilize word embeddings. Finally, we look at hybrid approaches that can accommodate relative amounts of semantic information into what is otherwise a purely lexical model.
From the retrieval point of view, all approaches fall into one of two categories: retrieval methods based on similarity in vector space and retrieval methods using domain-specific similarity scoring functions.

\subsection{Lexical}\label{sec:lexical}

\paragraph{Hand-crafted scoring function} Previous work has devised hand-crafted scoring functions targeted at retrieving intertextual relationships similar to those found in Bernard \cite{Forstall2015ModelingMatching}. The scoring function is used in an online retrieval system\footnote{
    The retrieval system can be accessed at the following URL: \href{http://tesserae.caset.buffalo.edu/}{http://tesserae.caset.buffalo.edu/}   
} and is defined by Eq.~\ref{eq:tes}:
\begin{equation}\label{eq:tes}
    T(s,t) = ln \left(\frac{\sum_{w\in (S \cap T)}\frac{1}{f_{(w, s)}} + \frac{1}{f_{(w, t)}}} {d_s + d_t}\right)
\end{equation}
where $f_{(w,d)}$ refers to the frequency of word $w$ in document $d$ and $d_d$ refers to the distance in tokens between the two most infrequent words in document $d$. Note that $T(s,t)$ is only defined for cases in which documents share at least 2 words, since otherwise the denominator cannot be computed. While this presents a clear disadvantage, it also lends itself to evaluation in a hybrid fashion with a complementary back-off model operating on passages with lower overlap. While originally $f_{(w,s)}$ is defined with respect to the query (or target) document, we observed such choice yielded poor performance (probably due to the small size of the documents), and, therefore, we use frequency estimates extracted from the respective document collections instead. We refer to this model as \texttt{Tesserae}.

\paragraph{BOW \& TfIdf} We include retrieval models based on a bag-of-words document representation (BOW) and cosine similarity for ranking. In a BOW space model, a document $d$ is represented by a vector where the $i_{th}$ entry represents the frequency of the $i_{th}$ word in $d$. Beyond word counts, it is customary to apply the Tf-Idf transformation, that targets the fact that the importance of a word for a document is also dependent on how specific it is to that document. Tf-Idf for the $i_{th}$ word is computed as the product of its frequency in $d$, denoted $Tf(w,d)$, and its inverse document frequency, $Idf(w,d)$, defined by Eq.~\ref{eq:idf}:
\begin{equation}\label{eq:idf}
    Idf(w,d) = log \left(\frac{|D|}{1 + |\{d \in D : w \in d\}|}\right)
\end{equation}
We refer to these retrieval models as \texttt{BOW} and \texttt{TfIdf}. Given document vector representations in some common space, we can compute their similarity score based on the cosine similarity between such vectors:

\begin{equation}
    cos(\overrightarrow{s}, \overrightarrow{t}) = \frac{\sum_i s_i t_i}{\sqrt{\sum_i s_i^2}\sqrt{\sum_i t_i^2}}
\end{equation}

\subsection{Semantic}

We define a number of semantic models based on distributional semantics and, in particular, word embeddings. We use \texttt{FastText} word embeddings \cite{bojanowski2017enriching} trained with default parameters on a large collection of Latin texts provided by \citep{bamman2011measuring}, which include 8.5GB of text of varying quality.\footnote{
    All the relevant materials are available at the following URL: \href{http://www.cs.cmu.edu/~dbamman/latin.html}{http://www.cs.cmu.edu/~dbamman/latin.html}. We also experimented with an LSI retrieval model \cite{Deerwester1990IndexingAnalysis}, similar to the one used by \citep{Scheirer2016TheMeaning}, but found it performed poorly on this dataset due to the small size of the documents in our dataset.
}


\paragraph{Sentence Embeddings} We use distributional semantic models based on the idea of computing a sentence embedding through a composition function operating over the individual embeddings of words in the sentence. The most basic composition function is averaging over the single word embeddings in the sentence \cite{DBLP:journals/corr/WietingBGL15a}. We can take into account the relative importance of words to a given sentence using the Tf-Idf transformation defined in Section~\ref{sec:lexical} and compute a Tf-Idf weighted average word embedding. We refer to these models as \texttt{BOW\textsubscript{emb}} and \texttt{TfIdf\textsubscript{emb}} respectively.

\paragraph{Word Mover's Distance} \texttt{WMD} is a metric based on the transportation problem known as Earth Mover's Distance but defined for documents over word embeddings. \texttt{WMD} has shown excellent performance in document retrieval tasks where semantics play an important role \cite{kusner2015word}. Intuitively, \texttt{WMD} is grounded on the idea of minimizing the amount of ``travel cost'' incurred in moving the word histogram of a document $s$ into the word histogram of $t$, where the ``travel distance'' between words $w_i$ and $w_j$ is given by their respective distance in the embedding space $cos(w_i,w_j)$. Formally, \texttt{WMD} is computed by finding a so-called flow matrix $T\in\mathbb{R}^{VxV}$---where $T_{ij}$ denotes how much of word $w_i$ in $s$ travels to word $w_j$ in $t$---such that $\sum_{i,j} T_{i,j} c(w_i,w_j)$ is minimized. Computing \texttt{WMD} involves solving a linear programming problem for which specialized solvers exist.\footnote{
    We use the implementation provided by the \texttt{pyemd} package \cite{laszuk2017pyemd}
}

\subsection{Hybrid}

We look into methods that are able to encompass both lexical and semantic information.

\paragraph{Tesserae + WMD as backoff model (\texttt{T+WMD})} Since \texttt{Tesserae} score is only defined for document pairs with at least 2 words in common, it can be easily combined with other models in a backoff fashion. In particular, we evaluate this setup using \texttt{WMD} as the backoff model since it proved to be the most efficient purely semantic model.\footnote{
    We note that for this retrieval setup to be used in practice $WMD$ and $Tesserae$ similarity scores must be transformed into a common scale. In the present paper, we assume an oracle on the lexical overlap with the relevant document and therefore the resulting numbers must be interpreted as an optimal score given perfect scaling.
}

\paragraph{Soft Cosine} 
A more principled approach to combining lexical and semantic information is based on the soft cosine similarity function, which was first introduced by \citep{sidorov2014soft} and has been recently used in a shared-task winning contribution by \citep{charlet2017simbow} for question semantic similarity. Soft cosine generalizes cosine similarity by considering not only how similar vectors $s$ and $t$ across feature $i$ but more generally across any given pair of features $i,j$. Soft cosine is defined by Eq.~\ref{eq:soft_cosine}:

\begin{equation}\label{eq:soft_cosine}
    soft\_cos(\overrightarrow{s},\overrightarrow{t}) = \frac{\sum_{i,j} S_{i,j} s_i t_j}{\sqrt{\sum_{i,j}S_{i,j}s_i s_j} \sqrt{\sum_{i,j}S_{i,j} t_i t_j}}
\end{equation}

with $S \in\mathbb{R}^{VxV}$ representing a matrix where $S_{i,j}$ expresses the similarity between the $i_{th}$ and the $j_{th}$ word in the vocabulary. It can be seen that soft cosine reduces to cosine when $S$ is taken to be the identity matrix.

Soft cosine is a flexible function since it lets us use any linguistic resource to estimate the similarity between words. For our purposes, matrix $S$ can be estimated on the basis of WordNet-based semantic relatedness measures or word embedding based semantic similarity estimates. 
More concretely, we define the following two models. 
\texttt{SC\textsubscript{wn}}, which uses a similarity function based on the size of the group of synonyms extracted from the Latin WordNet \cite{minozzi2010wordnet}:
$S_{i,j} = \frac{1}{|T_i \cap T_j|}$ where $T_i$ refers to the set of synonyms of the $i_{th}$ word. \texttt{SC\textsubscript{emb}} which exploits word embedding similarity $S_{i,j} = max(0, cos(\overrightarrow{w_i}, \overrightarrow{w_j})$ over embeddings $\overrightarrow{w_i}, \overrightarrow{w_j}$. All soft cosine-based retrieval models are applied on $TfIdf$ document representations. In agreement with previous research \cite{charlet2017simbow}, we boost the relative difference in similarity between the upper and lower quantiles of the similarity distribution by raising $S$ to the $n$th-power.\footnote{
    During development we found that raising $S$ to the 5th power yielded the best results across similarity functions in all cases.
}

\subsection{Evaluation}

Given a Bernardian reference as a query formulated by the annotators and the collection of Biblical candidate documents, all evaluated models produce a ranking. Using such a ranking, we evaluate retrieval performance over the set of queries $Q$ using Mean Reciprocal Rank\footnote{
    For clarity, we transform $MRR$ from the original $[0-1]$ range into the $[0-100]$ range. 
} %
($MRR$) \cite{Voorhees1999TheReport} defined in Eq.~\ref{eq:mrr}:
\begin{equation}\label{eq:mrr}
    MRR(Q) = \frac{1}{|Q|} \sum_{j=1}^{|Q|}\frac{1}{|R_j|}
\end{equation}

Additionally, we also report $Precision@K$---based on how often the system is expected to retrieve the relevant document within the first $k$ results---since it is a more interpretable measure from the point of view of the retrieval system user.

It must be noted that $P@K$ and $MRR$ are not suitable metrics to evaluate a text reuse detection system on unrestricted data, since, in fact, most naturally occurring text is not allusive. However, the focus of the present paper lies on the feasibility of allusive text detection, which we aim to elucidate on the basis of a pre-annotated dataset in which each query is guaranteed to match to a relevant document in the target collection. The results must therefore be interpreted taking into account the artificial situation, where the selected queries are already known to contain allusions and the question is how well different systems recognize the alluded verse.


\paragraph{Results}

\begin{table*}[ht]
\makebox[\textwidth]{
\scalebox{1}{

\setlength\tabcolsep{4pt} 
    \centering
    \begin{tabular}{rc|ccc|ccc|ccc}
\toprule
  & & \multicolumn{3}{c|}{Lexical} & \multicolumn{3}{c|}{Semantic} & \multicolumn{3}{c}{Hybrid} \\
  Metric & Lemma & \texttt{BOW} & \texttt{TfIdf} & \texttt{Tesserae} & \texttt{BOW\textsubscript{emb}} & \texttt{TfIdf\textsubscript{emb}} & \texttt{WMD} & \texttt{SC\textsubscript{wn}} & \texttt{SC\textsubscript{emb}} & \texttt{T+WMD} \\
\midrule 
\multirow{2}{*}{$MRR$}  & & 11.85 & 16.42 & 12.39 & 8.54  & 9.59  & 13.68 &       & 21.41 & 17.01 \\
           & \checkmark & 15.07 & 19.51 & 13.36 & 9.82  & 11.13 & 14.07 & 19.75 & \textbf{21.95} & 16.18 \\
 
\multirow{2}{*}{$P@10$} & & 20.16 & 30.59 & 19.20 & 15.50 & 18.11 & 24.14 &       & 37.31 & 29.22 \\
           & \checkmark & 27.30 & 34.43 & 25.79 & 16.87 & 20.99 & 25.38 & 35.25 & \textbf{39.64} & 31.14 \\
           
\multirow{2}{*}{$P@20$} & & 25.38 & 35.94 & 22.22 & 20.44 & 24.14 & 27.85 &       & 44.31 & 33.61 \\
           & \checkmark & 34.16 & 43.35 & 30.86 & 22.63 & 26.20 & 31.28 & 44.44 & \textbf{47.60} & 38.27 \\
\bottomrule
\end{tabular}
}
}
    \caption{Retrieval results for all considered models grouped by approach type. All models are evaluated with tokens and lemmas as input except for $SC_{wn}$ which requires lemmatized input. Overall best numbers per metric are shown in bold letters.}
    \label{tab:retrieval}
\end{table*}

As shown in Table~\ref{tab:retrieval}, the best model overall is $SC_{emb}$, achieving 21.95 $MRR$ and 47.60 $P@20$, closely followed by another soft cosine-based hybrid approach: $SC_{wn}$. Interestingly, a simple $TfIfd$ baseline over lemmatized input results in strong ranking performance, surpassing all other purely lexical -- including the hand-crafted $Tesserae$ -- and all purely semantic models. In agreement with general expectations, all models benefit from lemmatized input and $TfIdf$ transformation (both as input representation in purely lexical models and as a weighting scheme for the sentence embeddings in purely semantic approaches). $WMD$ outperforms any other purely semantic model, but as already pointed out, it compares negatively to the purely lexical $TfIdf$ baseline. The combination of $Tesserae$ with $WMD$ as back-off proves useful and outperforms both approaches in isolation, highlighting that they model complementary aspects of text reuse.

\begin{table}[t!]
    \centering
    \begin{tabular}{cc|ccc}
\toprule
& & \multicolumn{3}{|c}{Model} \\
Metric & Lemma & $SC_{emb}$ & $SC_{w2v}$ & $SC_{rnd}$ \\
\midrule
\multirow{2}{*}{$MRR$} & & 21.41 & 19.26 & 18.56 \\
          & \checkmark & 21.95 & 20.18 & 20.22 \\

\multirow{2}{*}{$P@10$} & & 37.31 & 33.33 & 31.28 \\
           & \checkmark & 39.64 & 36.35 & 35.67 \\

\multirow{2}{*}{$P@20$} & & 44.31 & 39.09 & 36.76 \\
           & \checkmark & 47.60 & 43.90 & 43.48 \\
\bottomrule
\end{tabular}
\caption{Comparison of soft cosine using \texttt{FastText} embeddings ($SC_{emb}$), \texttt{word2vec} embeddings ($SC_{w2v}$) and a random similarity baseline ($SC_{rnd}$).}
\label{tab:rnd}
\end{table}

In order to test the specific contribution of the similarity function used to estimate $S$, we compare results with soft cosine using a random similarity matrix ($S_{rnd}$) defined by Eq.~\ref{eq:rnd}:
set on\begin{equation}\label{eq:rnd}
S_{i,j} = 
     \begin{cases}
       1 &\quad i=j\\
       \thicksim \mathcal{N}(0.5, 0.05) &\quad otherwise \\ 
     \end{cases}
\end{equation}
We also investigate the effect of the word embedding algorithm by comparing to $SC_{emb}$ based on \texttt{word2vec} embeddings \cite{mikolov2013efficient}. As Table~\ref{tab:rnd} shows, \texttt{FastText} embeddings, an algorithm known to capture not just semantic but also morphological relations, yields strong improvements over \texttt{word2vec}. Moreover, a random approach produces strong results, only underperforming the \texttt{word2vec} model by a small margins, which questions the usefulness of the semantic relationships induced by \texttt{word2vec} for the present task.

Finally, we test the relative importance of the query segmentation to the retrieval of allusive text reuse. For this purpose, we evaluate our best model ($SC_{emb}$) on a version of the dataset in which the referencing text is segmented according to a window approach, selecting $n$ words around the anchor expression.

\begin{table}[t!]
    \centering
    \begin{tabular}{cc|ccc}
\toprule
& & \multicolumn{3}{|c}{Segmentation} \\
Metric & Lemma & Manual & Win-3 & Win-10 \\
\midrule
\multirow{2}{*}{$MRR$} & & 21.41 & 13.41 & 13.98 \\
          & \checkmark & 21.95 & 14.67 & 14.69 \\

\multirow{2}{*}{$P@10$} & & 37.31 & 25.79 & 25.10 \\
           & \checkmark & 39.64 & 25.93 & 26.47 \\

\multirow{2}{*}{$P@20$} & & 44.31 & 31.41 & 31.41 \\
           & \checkmark & 47.60 & 32.78 & 34.57 \\
\bottomrule
\end{tabular}
\caption{Comparison of best performing approach $SC_{emb}$ across different segmentation types: manual and automatic window of 3 (Win-3) and 10 (Win-10) tokens to each side of the anchor word.}
\label{tab:window}
\end{table}
As Table~\ref{tab:window} shows, results on manually segmented text are always significantly better than on automated segmentation. A window of 10-word around the anchor produces slightly better results than a 3-word window -- more closely matching the overall mean length of manually annotated queries. This indicates the importance of localizing the appropriate set of referential words in context, while avoiding the inclusion of confounding terms. In other words, both precision and recall matter to segmentation, an issue that has been observed previously \cite{Bamman2009DiscoveringTexts}.

\begin{figure}[ht]
    \centering
    \includegraphics[width=\linewidth]{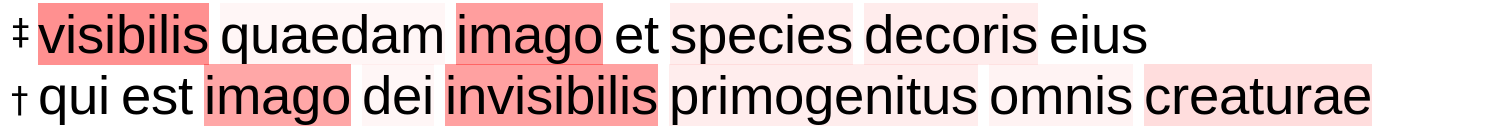}
    \caption{}
    \label{fig:visibilis}
\end{figure}

\begin{figure}[ht]
    \centering
    \includegraphics[width=\linewidth]{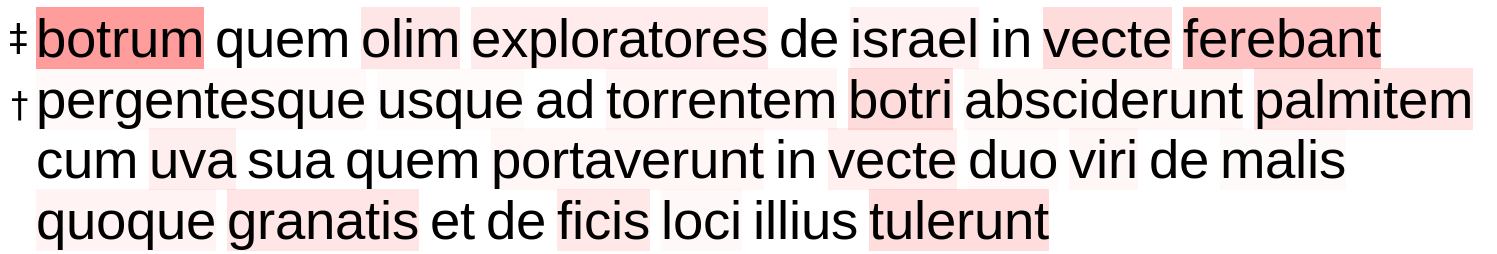}
    \caption{}
    \label{fig:botrum}
\end{figure}

\paragraph{Qualitative inspection} 
To appreciate the effect of the soft cosine using a semantic similarity matrix, it is worthwhile to inspect a hand-picked selection of items which were correctly retrieved by \texttt{SC\textsubscript{emb}} but not by \texttt{TfIdf}.\footnote{
    In the examples, we display the relative contribution made by each term in a sentence to the total similarity score (darker red implies higher contribution). Queries are preceded by a double dagger ($\ddagger$) and Bible references by a simple dagger ($\dagger$).
} In Fig~\ref{fig:visibilis}, the distributional approach adequately captures the antonymic relation between \emph{visibilis} ($\ddagger$) and \emph{invisibilis} ($\dagger$), which is reinforced by the synonymy between \emph{species} ($\ddagger$) and \emph{imago} ($\dagger$). Similar mechanisms seem at work in Fig~\ref{fig:botrum}, where the semantic similarity between vinery-related words increases the overall similarity score (\emph{botrus}, \emph{palmes}, \emph{uva}, \emph{granatus}).

\vspace{-0.1in}
\vspace{-0.1in}
\begin{figure}[ht]
    \centering
    \includegraphics[width=\linewidth]{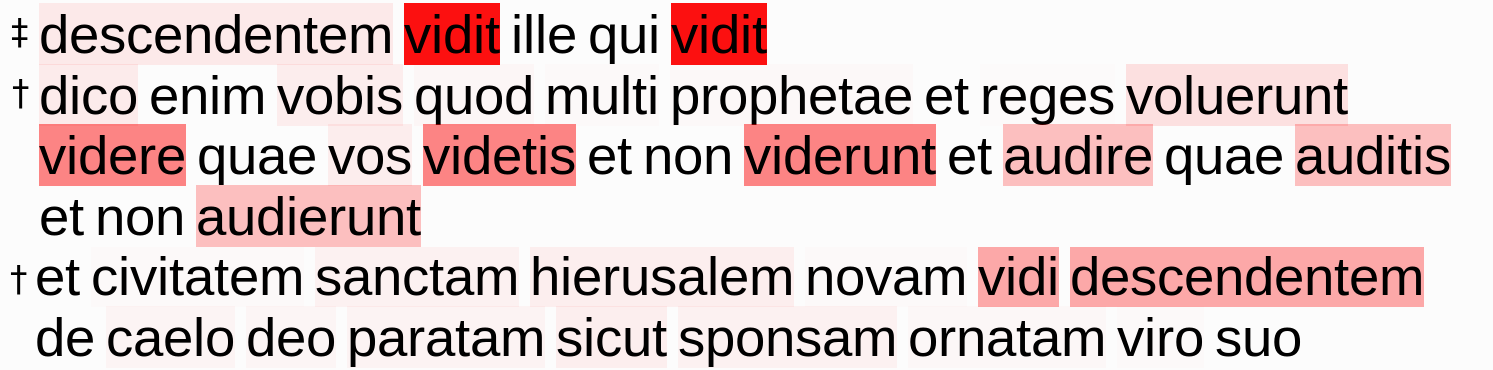}
    \caption{}
    \label{fig:descendem}
\end{figure}
\begin{figure}[ht]
    \centering
    \includegraphics[width=\linewidth]{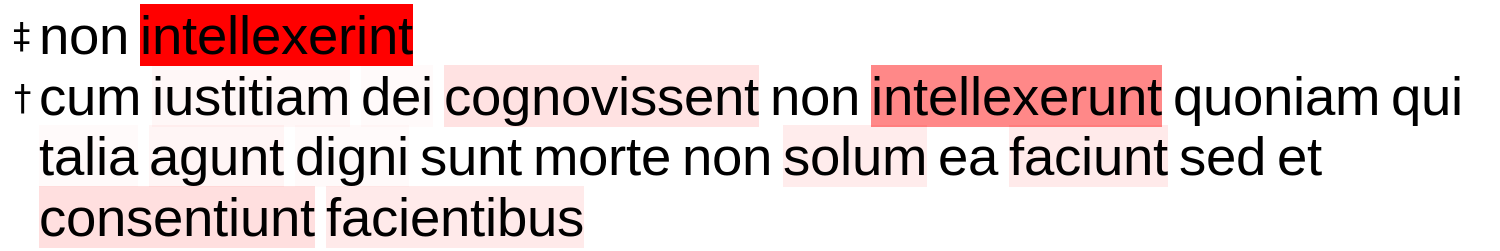}
    \caption{}
    \label{fig:intellexerint}
\end{figure}



Although the \texttt{SC} offers a welcome boost in retrieval performance, many errors remain. A first and frequent category are allusions that are simply hard to detect, even for human readers, often because they are very short or cryptic such as Fig~\ref{fig:intellexerint}, where despite increased semantic support---\emph{cognovissent} being synonymous with \emph{intellexerint}---the match is missed.

A second type of error occurs when less relevant candidates are pushed higher in the rank due to semantic reinforcements in the wrong direction. For example, in Fig~\ref{fig:descendem} we have a query together with a wrongly retrieved match (\emph{dico enim \ldots}) and the true, non retrieved reference (\emph{et civitatem \ldots}). We observe that due to the high similarity of redundantly repeated perception verbs (\emph{video}, \emph{audio}), the wrong match receives high similarity whereas the true reference remains at lower rank.

\section{Conclusions and Future Work}
Our experiments have highlighted the difficulties of automated allusion detection. Even assuming manually defined queries, the best performing model could only find the matching reference within the top 20 hits in less than half of the dataset. Moreover, the retrieval quality heavily drops when relying on windowing for query construction. This aspect calls for further research into the problem of automatic query construction for the detection of allusive reuse.

Across all our experiments, purely semantic models are consistently outperformed by a purely lexical TfIdf model. Similarly, lemmatization boosts the performance of nearly all models which also suggests that ensuring enough lexical overlap is still a crucial aspect of allusive reuse retrieval. A similar reasoning helps explaining the superiority of \texttt{FastText} over \texttt{word2vec} embeddings, since the former is better at capturing morphological relationships -- and lemma word embeddings suffer from data sparsity in the latter.

Overall, the hybrid models involving soft cosine show best performance, which indicates the effectiveness of such technique to incorporate semantics into BOW-based document retrieval and offers evidence that improvements in allusive reuse detection, however limited, can be gained from lexical semantics.

An interesting direction for future research is the application of soft cosine to text reuse detection across languages, leveraging current advances in multilingual word embeddings \cite{ammar2016massively} to extract multilingual word similarity matrices. Similarly, while the effect of adding semantic information from WordNet was less effective, it is still worth expanding the scope of semantic relationship beyond synonymy and exploring the usage of semantic similarity measures defined over WordNet \cite{budanitsky2001semantic}.


\section*{Acknowledgments} We are indebted to Laurence Mellerin for providing us with the dataset and to Dinah Wouters, Jeroen De Gussem, Jeroen Deploige and Wim Verbaal for their help in curating the dataset and providing invaluable feedback and discussions. 

\bibliography{references,extra}
\bibliographystyle{acl_natbib}

\end{document}